\def\year{2022}\relax
\documentclass[letterpaper]{article} 
\usepackage{aaai22}  
\usepackage{times}  
\usepackage{helvet}  
\usepackage{courier}  
\usepackage[hyphens]{url}  
\usepackage{graphicx} 
\usepackage{natbib}  
\usepackage{caption} 
\DeclareCaptionStyle{ruled}{labelfont=normalfont,labelsep=colon,strut=off} 
\usepackage{algorithm}
\usepackage{algorithmic}

\usepackage{amsmath,amsfonts,bm}

\def\eqref#1{equation~\ref{#1}}

\def\1{\bm{1}}

\DeclareMathAlphabet{\mathsfit}{\encodingdefault}{\sfdefault}{m}{sl}
\SetMathAlphabet{\mathsfit}{bold}{\encodingdefault}{\sfdefault}{bx}{n}

\usepackage[utf8]{inputenc} 
\usepackage[T1]{fontenc}    
\usepackage{hyperref}       
\usepackage{url}            
\usepackage{booktabs}       
\usepackage{amsfonts}       
\usepackage{nicefrac}       
\usepackage{microtype}      
\usepackage{xcolor}         
\usepackage{natbib}
\usepackage{hyperref}
\usepackage{url}
\usepackage{graphicx}
\setcitestyle{round}
\usepackage{xcolor}
\usepackage{wrapfig}
\usepackage{caption}
\usepackage{float}
\usepackage[position=bottom]{subfig}
\usepackage{balance}
\usepackage{newfloat}
\usepackage{listings}
\floatstyle{ruled}
\newfloat{listing}{tb}{lst}{}
\floatname{listing}{Listing}

\title{Conditional Synthetic Data Generation for Robust Machine Learning Applications\\ with Limited Pandemic Data}
\author {
    Hari Prasanna Das\textsuperscript{\rm 1},
    Ryan Tran\textsuperscript{\rm 1},
    Japjot Singh\textsuperscript{\rm 1},
    Xiangyu Yue\textsuperscript{\rm 1},
    Geoff Tison\textsuperscript{\rm 2},\\
    Alberto Sangiovanni-Vincentelli\textsuperscript{\rm 1},
    Costas J. Spanos\textsuperscript{\rm 1}
}
\affiliations {
    \textsuperscript{\rm 1} Department of Electrical Engineering and Computer Sciences, University of California, Berkeley\\
    \textsuperscript{\rm 2} Division of Cardiology, University of California, San Francisco (UCSF)\\
    \{hpdas,bobotran,calzoom,xyyue,alberto,spanos\}@berkeley.edu, geoff.tison@ucsf.edu
}

\usepackage{bibentry}

\begin{document}

\maketitle
\begin{abstract}
    \textbf{Background:} At the onset of a pandemic, such as COVID-19, data with proper labeling/attributes corresponding to the new disease might be unavailable or sparse. Machine Learning (ML) models trained with the available data, which is limited in quantity and poor in diversity, will often be biased and inaccurate. At the same time, ML algorithms designed to fight pandemics must have good performance and be developed in a time-sensitive manner. To tackle the challenges of limited data, and label scarcity in the available data, we propose generating conditional synthetic data, to be used alongside real data for developing robust ML models. \textbf{Methods:} We present a  hybrid model consisting of a conditional generative flow and a classifier for conditional synthetic data generation. The classifier decouples the feature representation for the condition, which is fed to the flow to extract the local noise. We generate synthetic data by manipulating the local noise with fixed conditional feature representation. We also propose a semi-supervised approach to generate synthetic samples in the absence of labels for a majority of the available data. \textbf{Results:} We performed conditional synthetic generation for chest computed tomography (CT) scans corresponding to normal, COVID-19, and pneumonia afflicted patients. We show that our method significantly outperforms existing models both on qualitative and quantitative performance, and our semi-supervised approach can efficiently synthesize conditional samples under label scarcity. As an example of downstream use of synthetic data, we show improvement in COVID-19 detection from CT scans with conditional synthetic data augmentation.
\end{abstract}
\begin{figure*}[t]
\centering
\includegraphics[width=0.83\textwidth]{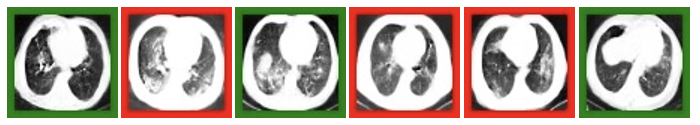}
\caption{Synthetic CT scans generated by our proposed model, with Non-COVID (normal and pneumonia cases, images with green border)/ COVID  (images with red border) as the condition.}
\label{fig:mainpage}
\end{figure*}
\section{Introduction}
The COVID-19 pandemic has created a public health crisis and continues to have a devastating impact on lives and healthcare systems worldwide. In the fight against this pandemic, a number of algorithms involving state-of-the-art machine learning techniques have been proposed. Data-based approaches have been used in a number of important tasks such as detection, mitigation, transmission modeling, decision on lockdown, reopening and related restrictions etc. For example, computer vision-based detection of COVID-19 from chest computed tomography (CT) images has been proposed as a supportive screening tool for COVID-19~\citep{gunraj2020covidnetct}, along with the primary diagnostic test of transcription polymerase chain reaction (RT-PCR). This is beneficial since obtaining definitive RT-PCR test results may take a lot of time in critical situations. Reinforcement learning based methods were also proposed to optimize mitigation policies that minimize the economic impact without overwhelming the hospital capacity~\citep{kompella2020reinforcement}.

The application of machine learning algorithms in healthcare depends upon ample availability of disease data along with their attributes/labels. At the beginning of a pandemic, data corresponding to the disease might be unavailable or sparse. Sparse data often have limited variation in several important factors relevant to disease detection such as age, underlying medical conditions etc. Class imbalance is another issue faced by machine learning algorithms when pandemic-disease related data is limited. For example, at the onset of COVID-19, the amount of CT scan images corresponding to COVID-19 were much less than those corresponding to other existing lung diseases (e.g. pneumonia). ML models fed with such class-imbalanced data could be biased and thus provide inaccurate results. Furthermore, the amount of data with proper labels among the available pandemic data might be minimal. This issue can arise because healthcare professionals and domain experts who can review and label the data are busy treating patients inflicted with the new disease, or also because of privacy concerns associated with medical data sharing. 

Concurrently, after a new disease has been discovered, the healthcare ML tools must rapidly adapt to the new disease in order to assist medical professionals diagnose and treat affected individuals as quickly as possible. Rapid actions are also expected in design of policy interventions that are based on insights from pandemic data. Another issue in development of machine learning algorithms for emerging pandemics is privacy. Development of solutions to pandemics at the scale of COVID-19 require collaborative research which in turn presses the need for open-sourced healthcare data. But, even if healthcare organizations wish to release relevant data, they are often restricted in the amount of data to be released due to legal, privacy and other concerns.

In this paper, we present a novel conditional synthetic data-generation method to augment the available pandemic data of interest. Our proposed method can also help organizations release synthetic versions of their actual data with similar behavior in a privacy-preserving manner. At the onset of a pandemic, when the availability of disease data is limited, our proposed model learns the distribution of available limited data and then generates conditional synthetic data that can be added to the existing data in order to improve the performance of machine learning algorithms. To tackle the challenge of label scarcity, we propose semi-supervised learning methods to leverage the small amount of labeled data and still generate qualitative synthetic samples. Our methods can enable healthcare ML tools to rapidly adapt to a pandemic.

We apply this method to generate conditional CT scan images corresponding to COVID cases, and conduct qualitative and quantitative tests to ensure that our model generates high-fidelity samples and is able to preserve the features corresponding to the condition (COVID/Non-COVID) in synthetic samples. As a downstream use of conditional synthetic data, we improve the performance of COVID-19 detectors based on CT scan data via synthetic data augmentation. Our results show that the proposed model is able to generate synthetic data that mimic the real data, and the generated samples can indeed be augmented with existing data in order to improve COVID-19 detection efficiency.
\section{Methodology}
\begin{figure*}[t]
    \centering
    \includegraphics[width=0.8\textwidth]{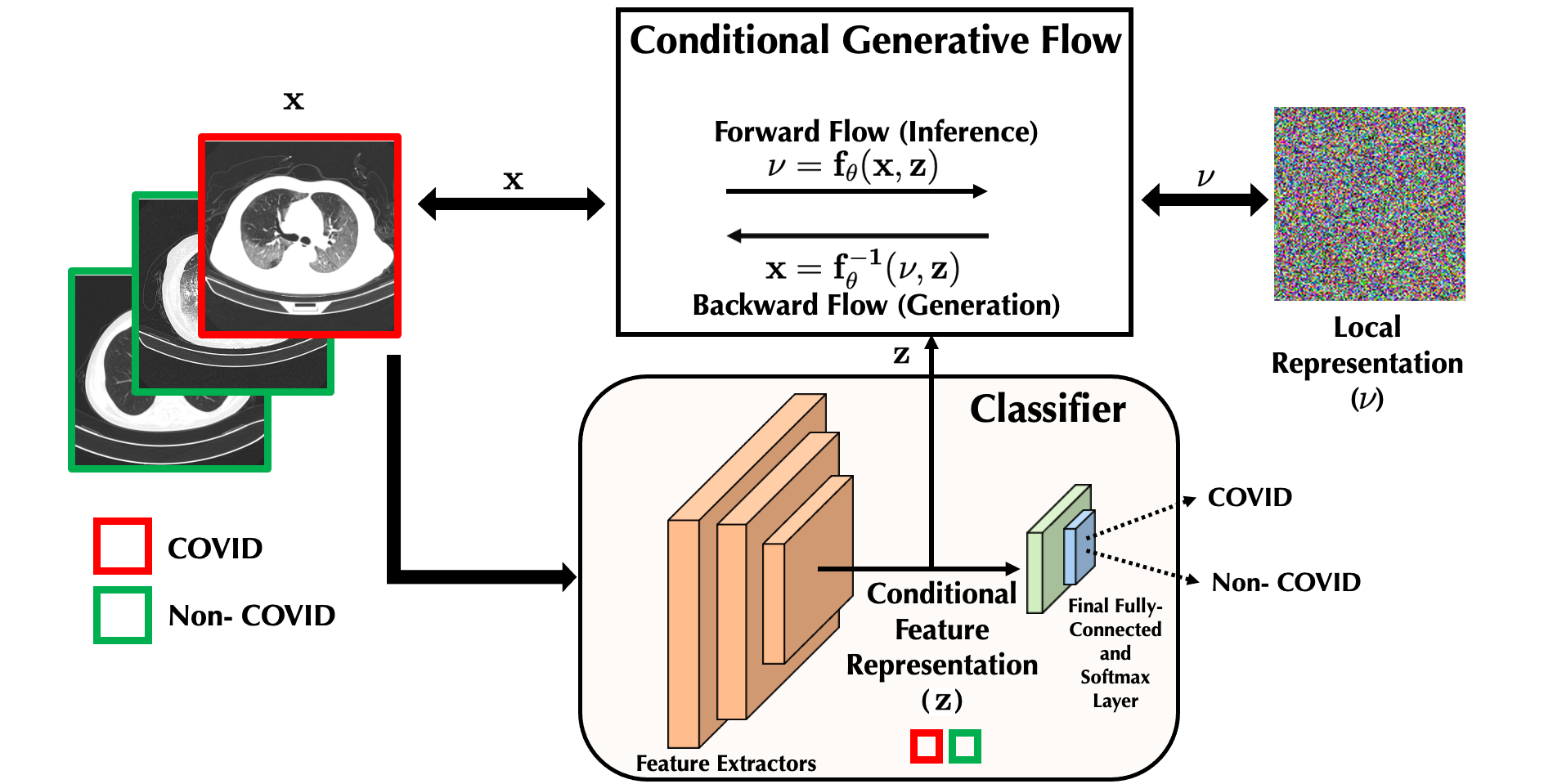}
    \caption{Illustration of the proposed conditional synthetic generation. (Best viewed in color)}
    \label{fig:schematic}
\end{figure*}
\begin{table*}[t]
  \centering
  \resizebox{\textwidth}{!}{%
  \begin{tabular}{p{8cm} p{9.2cm}}
    \toprule
    \multicolumn{1}{c}{\bf Inference Phase}  &\multicolumn{1}{c}{\bf Generation Phase}                   \\
    \midrule
    1. \textbf{(Classifier)} Train the COVID and Non-COVID&1. \textbf{(Classifier)} Corresponding to an input sample $x$, find its \\
\hspace{4mm}classifier.&\hspace{4mm}conditional feature representation $z$ using the trained classifier.\\
2. \textbf{(Flow)} For each input sample $x$,&2. \textbf{(Flow)} Sample a local representation $\Tilde{\nu}\sim\mathcal{N}(0,I)$.\\
\hspace{4mm} 2.1 Feed $x$ to the classifier and extract the conditional &3. \textbf{(Flow)} Get a synthetic sample $\Tilde{x} = f_{\theta}^{-1}(\Tilde{\nu},z)$.\\
\hspace{10mm}feature representation $z$ from its penultimate layer.&\\
\hspace{4mm} 2.2 Get the local representation as $\nu = f_{\theta}(x,z)$ &\\
\hspace{4mm} 2.3 Train the flow model with maximum-likelihood.\\
    \bottomrule
  \end{tabular}}
  \caption{Summary of steps for conditional inference and generation}
  \label{tab:inf_gen}
\end{table*}
We present a hybrid model consisting of a conditional generative flow and a classifier  for conditional synthetic generation. We also introduce a semi-supervised approach, to generate conditional synthetic samples when a few samples out of the whole dataset are labeled.

\subsection{COVID and Non-COVID Classifier}\label{meth:classifier}
Our model is characterized by the efficient decoupling of feature representations corresponding to the condition and the local noise. Suppose we have $N$ samples $\mathbf{X}$ with labels $Y$, with 2 possible classes, COVID/Non-COVID. We first train a classifier $C$ (consisting of a feature extractor network denoted by $g(\cdot)$, and a final fully-connected and softmax layer, denoted by $h(\cdot)$, i.e. $C(x) = h(g(x))$) to classify the input sample (which in our case are CT Scans) and associated labels as COVID and Non-COVID. Mathematically, this step solves the following minimization with backpropagation:
\begin{align}
    \min_{C} \mathcal{L}_{C}(\mathbf{X},Y) = -\mathbb{E}_{(x,y)\sim (\mathbf{X},Y)}\sum_{l=1}^{2}\left[\mathbb{I}_{[l=y]}\log C(x))\right]
\end{align}

By virtue of the training process, the classifier learns to discard local information and preserve the features necessary for classification (conditional information) towards the downstream layers. Once the classifier is trained, we freeze its parameters, and use it to extract the conditional (COVID/Non-COVID) feature representation $z = g(x)$ (as a vector without spatial characteristics) at the output of the feature extractor network for input image $x$. The dimension of $z$ is chosen such that $\dim(z)$ $<<$ $\dim(x)$.

\subsection{Conditional Generative Flow}\label{meth:cond_gen}
During the training phase for the flow model, the conditional feature representation $z$ is fed to the conditional generative flow. The flow model is trained using maximum-likelihood, transforming $x$ to its local representation $\nu$, i.e. 
\begin{align}
    f_{\theta}(x,z) = \nu\sim\mathcal{N}(0,I)
\end{align}
with $\nu$ having the same dimension as $x$ by the inherent design of flow models. We use the method introduced by \citet{ma2021decoupling} to incorporate the conditional input $z$ in flow model. Coupling layers in affine flow models have scale $(s(\cdot))$ and shift $(b(\cdot))$ networks~\citep{dinh2017density,das2019dimensionality}, which are fed with inputs after splitting, and their outputs are concatenated before passing on to the next layer. We incorporate the conditional information $z$ in the scale and shift networks. Mathematically, (with $x$ as the input, $D$ as input dimension, $d$ as the split size,and $y$ as output of the layer),
\begin{align*}
    x_{1:d},x_{d+1:D} &= \text{split}(x)\\
    y_{1:d} &= x_{1:d}\\
    y_{d+1:D} &= s(x_{1:d},z)\odot x_{d+1:D} + b(x_{1:d},z)\\
    y &= \text{concat}(y_{1:d},y_{d+1:D})
\end{align*}
Since flow models are bijective mappings, the exact $x$ can be reconstructed by the inverse flow with $z$ and $\nu$ as inputs. During the generation phase, for an input sample $x$, we compute the conditional feature representation $z$. Keeping the conditional feature representation the same, we sample a new local representation $\Tilde{\nu}$, and generate a conditional synthetic sample $\Tilde{x}$, i.e.

\begin{align}
    \Tilde{\nu}\in\mathcal{N}(0,I), \;\;\Tilde{x} = f_{\theta}^{-1}(\Tilde{\nu},z)
\end{align}
Here, $\Tilde{x}$ has the same conditional (COVID/Non-COVID) features as $x$ , but has a different local representation. An illustration of the proposed model is provided in Fig.~\ref{fig:schematic} and the steps for the inference and generation phases are summarized in Table~\ref{tab:inf_gen}.

\subsection{Semi-supervised Learning for Conditional Synthetic Generation under Label Scarcity}\label{meth:semi_supervised}
In reality, often a small amount of the already limited pandemic data available are labeled. Consider this case when a few of the datapoints are labeled, denoted by $\{\mathbf{X}^l,Y^l\}$. The rest of the data (unlabeled) is denoted by $\mathbf{X}^u$. To generate conditional synthetic samples under such label scarce situations, we propose a semi-supervised method to modify the classifier design process, in order to effectively decouple the feature representations corresponding to the conditions.

\subsection{Label learning algorithm} We first design a label learning algorithm to assign presumptive labels $\Tilde{Y}^l$ to the unlabeled samples $\mathbf{X}^u$. Assuming $k_i$ labeled samples are available for class $i$, we train the classifier network using the labeled samples only and compute in the embedded ($z$) space (1) the centroid vector $c_i$ for each class and (2) a similarity metric between each unlabeled target sample $x^u \in \mathbf{X}^u$ and the specific centroid. Depending on the dimension of the transformed feature space, this similarity metric can simply be a Gaussian kernel to capture local similarity~\citep{van2008visualizing}, or the inverse of
Wasserstein distance~\citep{shen2018wasserstein} for better generalization with complex networks. 

\subsection{Semi-supervised model training} 
Ideally, the semi-supervised scheme should be able to (1)
identify the correct labels of unlabeled target samples, and (2) update the classifier with the additional information. We establish an alternating approach that recursively performs (1) fixing the feature mapping $g$ and propagating presumptive labels using a greedy assignment, i.e., an unlabeled sample is presumed to have the same label to its closest centroid, and (2) updating the feature mapping (the classifier) as supervised learning by treating the presumptive labels as true labels.

The proposed greedy propagation, intuitively simple and
practically easy to implement, in fact has theoretical guarantees since the entropy objective is approximately submodular when the feature mapping is fixed. Please refer to~\citep{zhou2016causal} for a detailed theoretical analysis. The above is conducted alternately until the
convergence of the feature mapping and presumptive label
assignment. In practice, the convergence
is usually achieved in a few iterations. Once the classifier has been trained with this semi-supervised approach, the conditional generative flow training is performed as specified before in conditional generation section.
\begin{table*}[t]
\begin{minipage}[b]{0.53\textwidth}
\centering
\includegraphics[width=\textwidth]{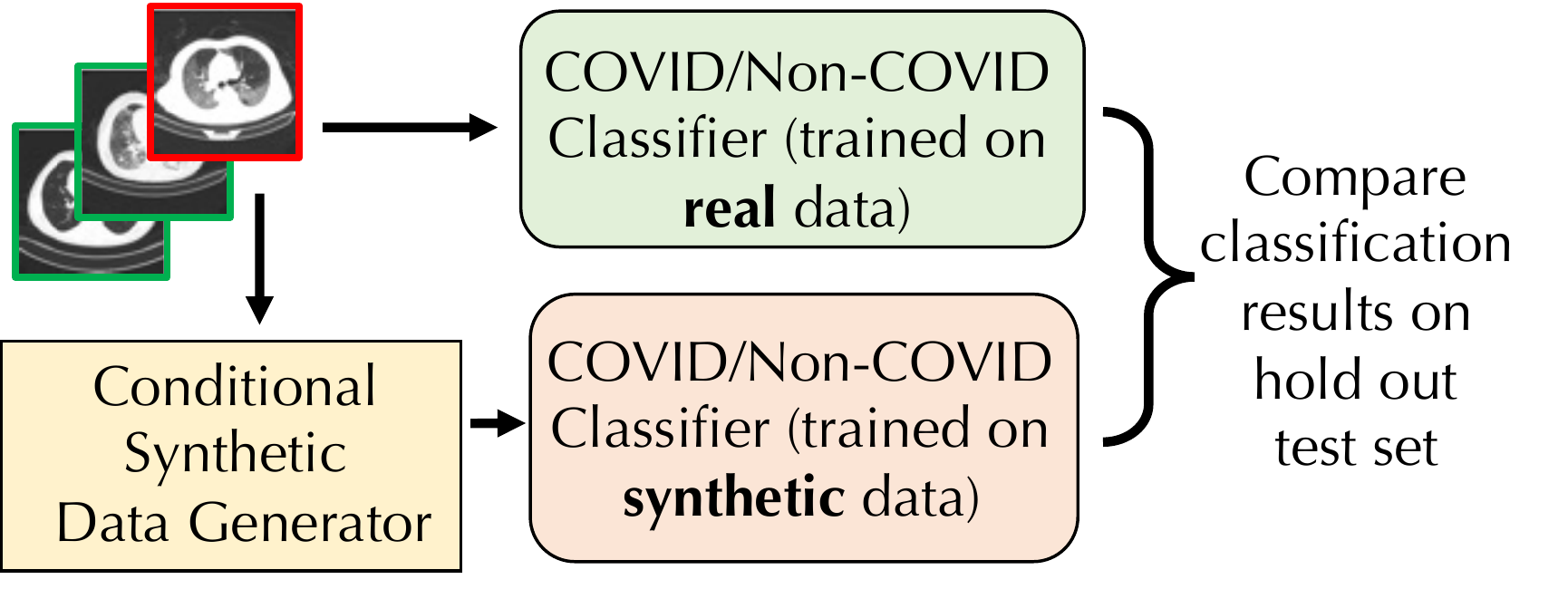}
\captionof{figure}{Illustration of quantitative testing procedure for conditional synthetic generation.}
\label{fig:test1}
\end{minipage}\hfill
\begin{minipage}[b]{0.45\textwidth}
\centering
\resizebox{0.5\textwidth}{!}{%
\begin{tabular}{p{2.25cm}p{2cm}p{2cm}p{2cm}p{2cm}p{2cm}}
\toprule
\multicolumn{1}{c}{\textbf{Model}} & \multicolumn{1}{c}{\textbf{FID}} \\ \midrule
\citet{ma2021decoupling}   & \multicolumn{1}{c}{$0.2504$}  \\
ACGAN  & \multicolumn{1}{c}{$0.0986$}           \\
CAGlow & \multicolumn{1}{c}{$0.0483$}          \\
Ours   & \multicolumn{1}{c}{$\mathbf{0.0077}$}                           \\\bottomrule

\end{tabular}}
\caption{Qualitative (Fr\'echet Information Distance) scores for synthetic data generated by various models (the lower the better).}
\label{tab:fid}
\end{minipage}
\begin{minipage}[b]{1\textwidth}
\centering
\includegraphics[width=\textwidth]{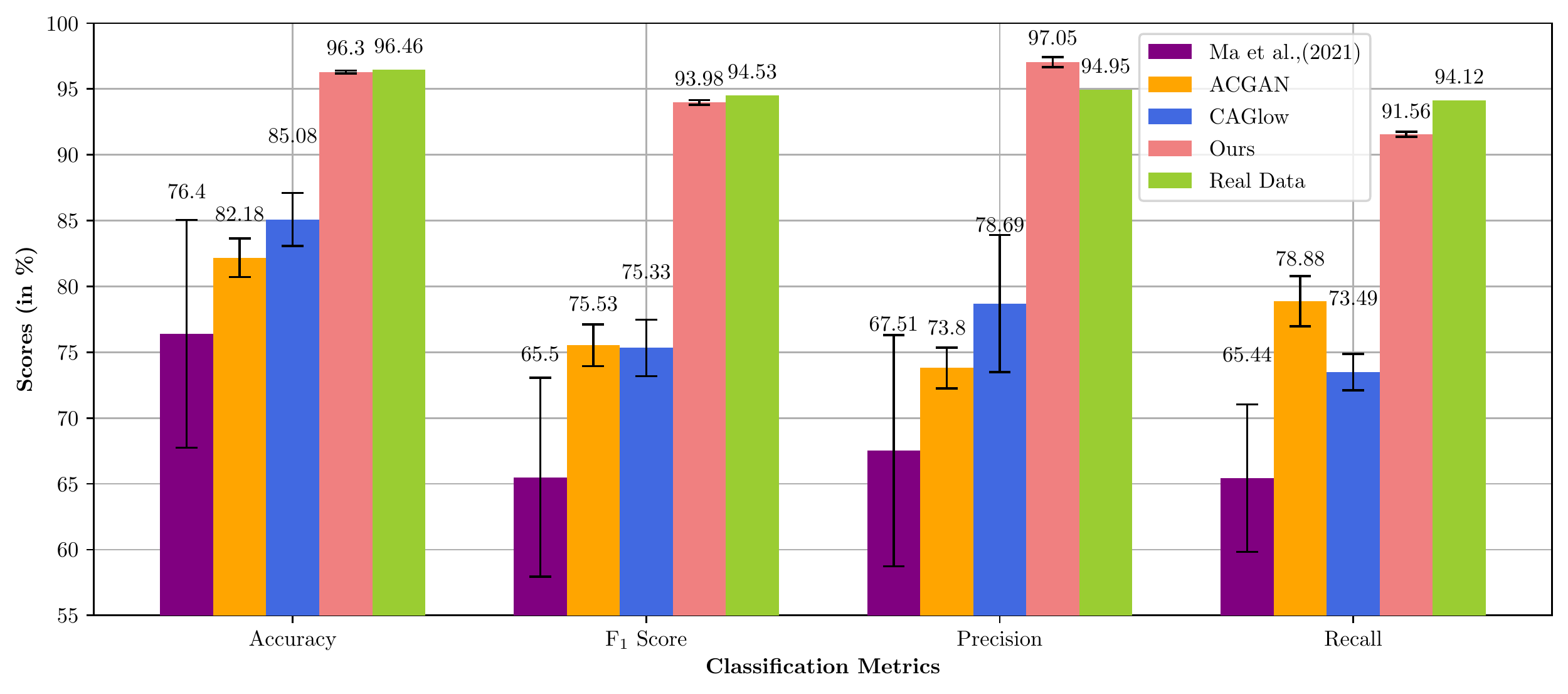}
\captionof{figure}{Classification metrics for classifiers trained on synthetic data generated by various models. The error bars indicate the variation in classifier performance when the synthetic datasets used to train them were generated multiple times with different seeds. Real data classifier does not involve multiple synthetic data generation, so its error bars are not included.}
\label{fig:results_main}
\end{minipage}\hfill
\end{table*}
\vspace{14mm}
\section{Experiments}
\subsection{Data Collection and Pre-processing}\label{sec:data}
We conduct experiments on chest CT scan data based on the COVIDx CT-1 dataset \citep{gunraj2020covidnetct}, publicly available in Kaggle\footnote{The CT scan dataset can be accessed at \url{www.kaggle.com/hgunraj/covidxct/version/1}}.

\textbf{CT Scan Data:} The dataset consists of 45,758 images corresponding to healthy individuals, 36,856 images corresponding to individuals afflicted with common pneumonia, and 21,395 images corresponding to individuals with COVID-19.

\textbf{Pre-processing:} We combine the images in the Normal and Pneumonia classes into a single Non-COVID class. We use the train, validation, and test splits defined by the official annotation files. In addition to class labels, the annotations include bounding boxes for the lungs region in the whole CT scans image. We crop the images as per the bounding box and resize them to $64\times 64$.

\textbf{Hyperparameters, Network Design and Computation used:} Please refer to Appendix.
\begin{figure*}[t]
    \centering
    \includegraphics[width=0.9\textwidth]{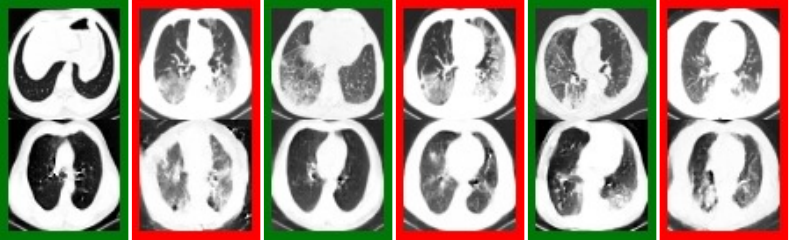}
    \caption{Original and generated synthetic CT scan samples. The top row consists of original samples, and corresponding image in the bottom row is the synthetic sample obtained by preserving the original conditional feature representation, and varying the local noise. Image pairs with a red border: COVID samples, and a green border: Non-COVID samples.}
    \label{fig:syn_gen}
\end{figure*}
\subsection{Testing Procedure}
We performed both quantitative and qualitative testing for conditional synthetic data generation by our model. A test set is held out from the real dataset to be used for quantitative testing. We then compare the classification performance (COVID/Non-COVID) on this test set for a classifier trained on real data vs a classifier trained on the generated synthetic data. This testing procedure is illustrated in Fig.~\ref{fig:test1}.
Since the datasets are imbalanced, we report the precision, recall and macro-F$_1$ score (together referred to as classification metrics) along with the accuracy. For more information on the metrics, please refer to~\citet{classification_metrics}. Closeness of the classification metrics of classifiers trained on synthetic and real data indicates an efficient design of the conditional synthetic generator. To evaluate the quality of generated samples, we report the Fr\'echet Inception Distance (FID) \citep{heusel2018gans} for the synthetic samples. 
For FID calculation, we use the embeddings from our classifier trained using real data, in place of the official inception network~\citep{szegedy2014going}, since the latter is not trained on medical imaging data.

\subsection{Results: Conditional Synthetic Data Generation}

The classification results for a classifier trained on the real data vs a classifier trained on purely conditional synthetic data, and tested on a hold-out set of real data, is given in Fig.~\ref{fig:results_main}. Across the existing methods for conditional synthetic generation, the classifier trained with synthetic data from our proposed model has the closest accuracy, F$_1$ score, precision and recall to that of the classifier trained on real data. This shows the capability of our method to generate synthetic samples with a distribution that closely matches the real conditional data distribution. The qualitative results (FID scores) for synthetic data generated by various models are tabulated in Table~\ref{tab:fid}. The FID scores for our model is the lowest among all models, demonstrating that the quality of the generated samples closely matches the real ones.

It is worth noting that the accuracy/F$_1$ score of the classifier trained with synthetic data generated by \citet{ma2020decoupling} is much smaller as compared to those by other models, not to mention the classifier trained on real data. This can be justified from the fact that \citet{ma2020decoupling} relies on an unsupervised method of decoupling global and local information. But for conditional synthetic generation applications, such as the one presented in this paper, the model needs information on what the model designer/ domain experts consider as the conditional information (COVID/Non-COVID in our case). ACGAN and CAGlow have different generators, but both include an auxiliary supervision signal to conditionally guide the generation process. Hence, performance of classifiers trained on synthetic data generated by them are close. We encode the conditions using feature extractors to feed to the generator, leading to state-of-the-art results.

The original samples along with the synthetic samples generated by preserving original conditional feature  representation and a different local noise for CT scans are shown in Fig.~\ref{fig:syn_gen}. The characteristic features for COVID CT scan samples, i.e., ground-glass opacity are well preserved in the synthetic samples. At the same time, the non-conditional local features, e.g. axial plane position for CT scans are considered as local noise. Since original samples for normal and pneumonia cases are merged together to form a single Non-COVID class, sometimes the corresponding synthetic image for a normal sample is a sample with pneumonia characteristics and vice-versa. This occurs since the conditional model learns to treat them as local information. The ability to decouple the feature representations for given conditions from other information in the data, as exhibited by our model, should be considered the strength of an effective conditional generative model.  

\subsection{Results: Conditional Synthetic Generation under Label Scarcity}
Previously, we proposed a semi-supervised learning approach to efficiently generate conditional synthetic samples when the number of samples labeled out of the available pandemic data is less. To test our approach, we retained the assigned label (COVID/Non-COVID) for a few samples, and discarded the label for rest of the samples. The amount of labeled samples was varied from 20 samples to 50 samples to 0.5\%, 1\%, and 5\% of the total training data. The ratio between COVID and Non-COVID samples was maintained among the labeled samples. We conducted the presumptive-labeling and classifier training in an iterative manner, and followed by this, trained the conditional generative flow using the conditional feature embeddings obtained using the feature extractors. We then generated conditional synthetic data using the above trained generative model. To show the robustness of our method, we perform bootstrapping on the test set and repeat our experiments using different sets of labeled samples from the training data. For each model, we also evaluated on multiple synthetic sets generated using random seeds. The results of classification models trained on the synthetic data under different bootstraps and seeds is given in Table~\ref{tab:semi_supervised}. 
\begin{table*}[t]
\begin{center}
\subfloat[With different sets of labeled samples and test set bootstrapping\label{tab:semi_supervised_bootstrap}]{
\resizebox{0.8\textwidth}{!}{%
\begin{tabular}{p{4cm}|p{2.5cm}|p{2.5cm}|p{2.5cm}|p{2.5cm}}
\toprule
\multicolumn{1}{c|}{\textbf{Amount of labeled data}} & \multicolumn{1}{c|}{\textbf{Accuracy (\%)}} & \multicolumn{1}{c|}{\textbf{F$_1$ Score (\%)}} & \multicolumn{1}{c|}{\textbf{Precision (\%)}} & \multicolumn{1}{c}{\textbf{Recall (\%)}}\\
\midrule
\multicolumn{1}{c|}{20 samples} & \multicolumn{1}{c|}{$84.84\pm 2.91$} & \multicolumn{1}{c|}{$76.32\pm 5.24$} & \multicolumn{1}{c|}{$77.15\pm 4.87$} & \multicolumn{1}{c}{$76.35\pm 5.87$}\\
\multicolumn{1}{c|}{50 samples} & \multicolumn{1}{c|}{$90.87\pm 1.31$} & \multicolumn{1}{c|}{$85.86\pm 1.73$} & \multicolumn{1}{c|}{$86.48\pm 2.68$} & \multicolumn{1}{c}{$85.43\pm 1.32$}\\
\multicolumn{1}{c|}{0.5\% of training samples} & \multicolumn{1}{c|}{$93.90\pm 0.46$} & \multicolumn{1}{c|}{$90.49\pm 0.61$} & \multicolumn{1}{c|}{$91.30\pm 1.28$} & \multicolumn{1}{c}{$89.8\pm 0.68$}\\
\multicolumn{1}{c|}{1\% of training samples} & \multicolumn{1}{c|}{$95.06\pm 0.49$} & \multicolumn{1}{c|}{$92.14\pm 0.69$} & \multicolumn{1}{c|}{$93.94\pm 1.30$} & \multicolumn{1}{c}{$90.62\pm 0.48$}\\
\multicolumn{1}{c|}{5\% of training samples} & \multicolumn{1}{c|}{$95.80\pm 0.20$} & \multicolumn{1}{c|}{$93.24\pm 0.28$} & \multicolumn{1}{c|}{$95.09\pm 0.84$} & \multicolumn{1}{c}{$91.23\pm 0.50$}\\
\midrule
\multicolumn{1}{c|}{100\% of training samples} & \multicolumn{1}{c|}{$96.30\pm 0.11$} & \multicolumn{1}{c|}{$93.98\pm 0.17$} & \multicolumn{1}{c|}{$97.05\pm 0.38$} & \multicolumn{1}{c}{$91.56\pm 0.20$}\\
\bottomrule
\end{tabular}}}\\
\subfloat[With multiple synthetic sets generated using random seeds]{
\resizebox{0.8\textwidth}{!}{%
\begin{tabular}{p{4cm}|p{2.5cm}|p{2.5cm}|p{2.5cm}|p{2.5cm}}
\toprule
\multicolumn{1}{c|}{\textbf{Amount of labeled data}} & \multicolumn{1}{c|}{\textbf{Accuracy (\%)}} & \multicolumn{1}{c|}{\textbf{F$_1$ Score (\%)}} & \multicolumn{1}{c|}{\textbf{Precision (\%)}} & \multicolumn{1}{c}{\textbf{Recall (\%)}}\\
\midrule
\multicolumn{1}{c|}{20 samples} & \multicolumn{1}{c|}{$85.70\pm 0.32$} & \multicolumn{1}{c|}{$78.65\pm 0.65$} & \multicolumn{1}{c|}{$77.96\pm 0.49$} & \multicolumn{1}{c}{$79.48\pm 1.18$}\\
\multicolumn{1}{c|}{50 samples} & \multicolumn{1}{c|}{$90.74\pm 0.77$} & \multicolumn{1}{c|}{$85.27\pm 0.88$} & \multicolumn{1}{c|}{$86.93\pm 2.03$} & \multicolumn{1}{c}{$83.98\pm 0.68$}\\
\multicolumn{1}{c|}{0.5\% of training samples} & \multicolumn{1}{c|}{$94.66\pm 0.86$} & \multicolumn{1}{c|}{$91.41\pm 1.27$} & \multicolumn{1}{c|}{$93.93\pm 2.02$} & \multicolumn{1}{c}{$89.39\pm 0.80$}\\
\multicolumn{1}{c|}{1\% of training samples} & \multicolumn{1}{c|}{$95.04\pm 0.32$} & \multicolumn{1}{c|}{$92.00\pm 0.47$} & \multicolumn{1}{c|}{$94.53\pm 0.88$} & \multicolumn{1}{c}{$89.96\pm 0.42$}\\
\multicolumn{1}{c|}{5\% of training samples} & \multicolumn{1}{c|}{$95.62\pm 0.21$} & \multicolumn{1}{c|}{$92.95\pm 0.28$} & \multicolumn{1}{c|}{$95.33\pm 0.77$} & \multicolumn{1}{c}{$90.99\pm 0.18$}\\
\midrule
\multicolumn{1}{c|}{100\% of training samples} & \multicolumn{1}{c|}{$96.30\pm 0.11$} & \multicolumn{1}{c|}{$93.98\pm 0.17$} & \multicolumn{1}{c|}{$97.05\pm 0.38$} & \multicolumn{1}{c}{$91.56\pm 0.20$}\\
\bottomrule
\end{tabular}}}
\end{center}
\caption{Results for classifiers trained on synthetic data generated by models that are developed using a few labeled data.}
\label{tab:semi_supervised}
\end{table*}

As is apparent from the table, using even a few labeled samples, our method is able to achieve results on par with the case when all the labels are available. This further reinforces the strength of our approach in generating conditional synthetic data to rapidly adapt ML models to a new pandemic at its onset, when there is scarcity of such labels. As expected, at lower levels of labeled data, the uncertainity associated with synthetic data generation is high, as is apparent from Table~\ref{tab:semi_supervised_bootstrap}, which dies down as we increase the labeled data amount. The uncertainity associated with classification models trained on synthetic set generated by our model using different seeds is low. Both the above observations establish the robustness of proposed method. 

An important point to note here is that the closeness of results obtained by utilizing 5\% of labels as compared to using 100\% of labels do not denounce the importance of the rest 95\% of labels. In healthcare, improvement of even 1\% of accuracy/F$_1$ score corresponds to a significant number of samples classified accurately, important especially during a pandemic. Thus, our proposed semi-supervised approach should be considered as a remedy for cases when labels are scarce, not as an alternative to fully-supervised approach.
\begin{figure}
    \centering
    \includegraphics[width=0.49\textwidth]{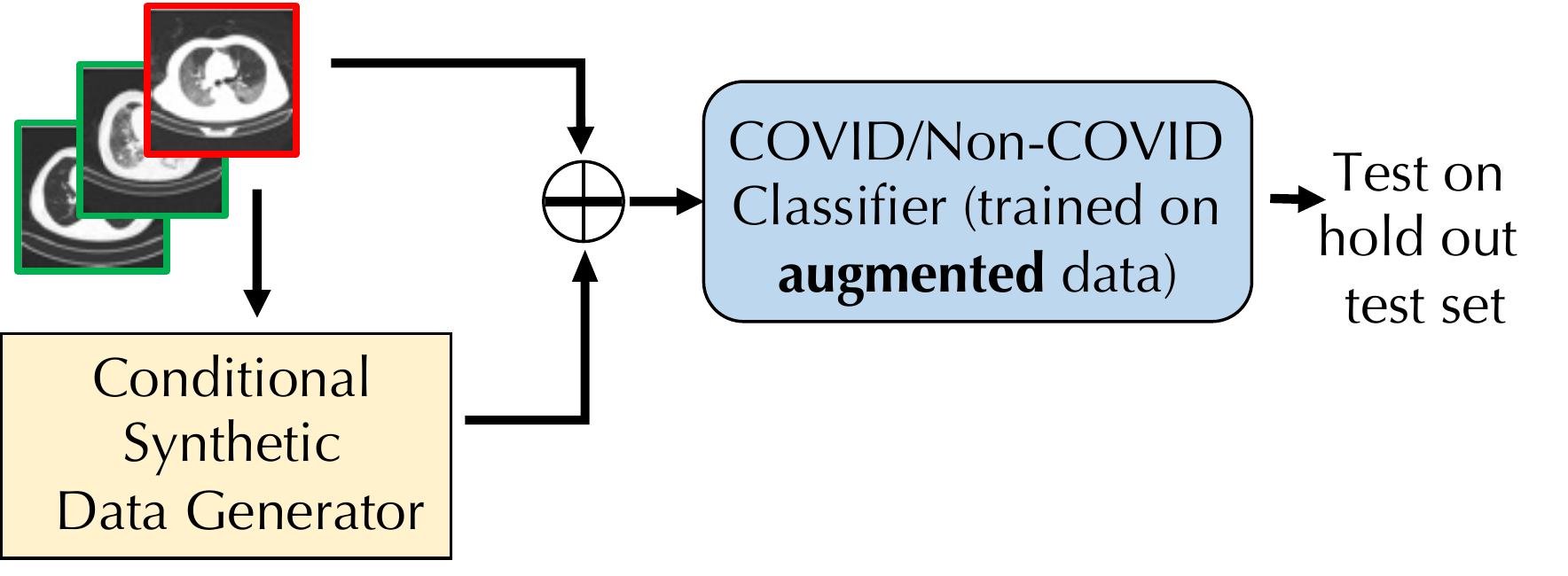}
    \caption{Illustration of synthetic data augmentation and testing process. Improvement in performance of classifiers trained on augmented data as compared to that trained on original training data is a step towards robust COVID-19 detection.}
    \label{fig:test2}
    \centering
    \includegraphics[width = 0.49
    \textwidth]{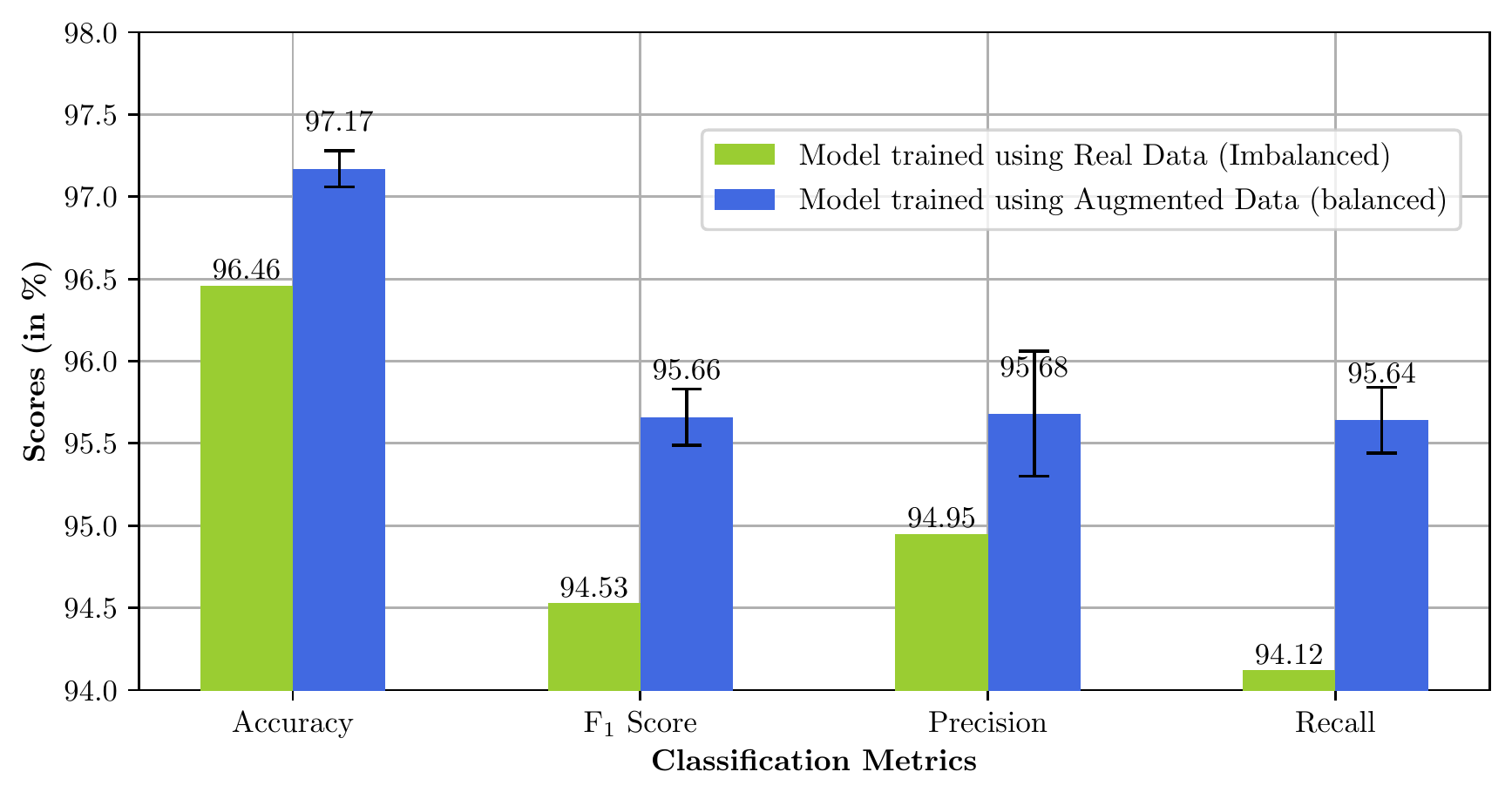}
    \caption{Classification results for models trained using real data (with class imbalance) vs augmented data (class-balanced). The real data (having $\sim 20\%$ of COVID samples) was augmented with synthetically generated COVID samples using the proposed model for class balancing.}
    \label{fig:results_aug}
\end{figure}
\subsection{Example Use of Synthetic Data: Robust Detection of COVID-19 via Data Augmentation}
Generated synthetic data can be utilized in a number of downstream tasks. We conduct experiments on one of the tasks: robust detection of COVID-19 via synthetic data augmentation.  The training data is inherently highly class-imbalanced, with limited samples of COVID and abundant samples for pneumonia and normal cases.
To design a robust COVID-19 detection mechanism under such class imbalance scenario, we augment the training data with synthetic COVID  samples generated using the proposed model to increase the \% of COVID samples and balance the dataset. The augmentation process and the testing procedure is illustrated in Fig.~\ref{fig:test2}. The classification metrics for classifiers trained on the augmented training data are given in Fig~\ref{fig:results_aug}. 

Examining the classification results, the classifier trained on augmented training data have better performance as compared to classifiers trained only on limited real training data for all augmentation levels. Note that even slight improvement in the recall score translates to numerous samples classified correctly (e.g. 1\% improvement in recall for CT scan corresponds to ~200 more correctly classified samples), leading to better diagnosis leading to accurate and timely treatment.



\section{Related Work}
In the field of healthcare, synthetic data generation has been proposed to expand the diversity and amount of the existing training data, often to improve the robustness of machine learning models. \citet{ghorbani2019dermgan} propose a generative adversarial network (GAN)-based synthetic data generator to improve the diversity and the amount of skin lesion images. \citet{kohlberger2019whole} synthesize pathology images for cancer with realistic out-of-focus characteristics to evaluate general pathology images for focus quality issues. \citet{Han841619} propose synthetic generation to produce high-resolution artificial radiographs. 
In the space of combating COVID-19, ~\citet{Bannur2020synthetic} propose a method of strengthening the COVID-19 forecasts from compartmental models by using short term predictions from a curve fitting approach as synthetic data. Similarly, ~\citet{waheed2020covidgan} and ~\citet{jiang2020covid} propose a conditional GAN-based generator for synthetic chest X-ray/CT scan data generation and augmentation for robust COVID-19 detection. Above works do not focus on the case where data with proper labels might be unavailable or sparsely available, whereas we tackle this challenge using a semi-supervised approach. We also show the robustness achieved using our model via experiments with several bootstrapping methods.

In the area of conditional generation, a hybrid flow and a GAN-based model have been proposed in CAGlow~\citep{liu2019conditional}. In general, GAN-based methods are known to be hard to train~\citep{salimans2016improved} and do not provide a latent embedding suitable for feature manipulations~\citep{kingma2018glow}. In contrast, we proposed a conditional generation method with efficient decoupling of the conditional information and local noise over an embedding space, along with a flow based generator, which recently have proved efficient in synthetic data generation~\citep{ho2019flow,das2021cdcgen}. We compared results for our proposed method over CAGlow and ACGAN for synthetic COVID CT scan generation, and showed improved results.

Decoupling of global and local representation for synthetic generation has been proposed in~\citet{ma2021decoupling}, where the global information is decoupled using a Variational AutoEncoder (VAE)~\citep{kingma2014autoencoding}. For conditional synthetic generation, it is necessary that the feature representations salient to the given conditions (COVID/Non-COVID) are decoupled from local noise, which is not guaranteed while extracting the same using a VAE. By employing a classifier network for the same, we ensure the relevant conditional information is not lost into the local noise. 

Semi-supervised learning based approaches to enhance classification models has been prominent in domain adaptation tasks, where except for a few samples, knowledge about the labels are generally unavailable in the target domain. A number of domain adaptation models, such as FADA~\citep{motiian2017fewshot}, \citet{teshima2020few}, \citet{zou2019consensus}, \citet{zhao2021domain} etc. employ few-shot learning approach, leveraging the few labeled data available to make the model efficient. In the space of healthcare, semi-supervised learning approaches have been used for skin disease identification from limited labeled samples in ~\citet{mahajan2020meta}, to enhance X-ray classification in \citet{rajan2021self} and in COVID-19 detection from scarce chest x-ray image data in ~\citet{jadon2021covid}. We proposed the use of semi-supervised learning in the space of synthetic data generation, to adapt our proposed generative model to label scarce scenarios, common at the onset of a pandemic.

\section{Discussions}\label{sec:discussions}
We presented a novel conditional synthetic generative model aimed at multiplying the samples of interest at the onset of a pandemic. We conducted extensive experiments on chest CT scan dataset to show the efficacy of the proposed model, and improvements in COVID-19 detection performance  achieved via synthetic data augmentation. We also proposed and experimented on a semi-supervised learning approach to efficiently generate conditional synthetic data under label scarce conditions. One of the limitations of our proposed method is that it does not exert selective control over the choice local noise, which can sometimes contain information for important interactions in the data, e.g., in our experiments, we extract conditional information salient to COVID/Non-COVID, whereas the information corresponding to everything else, such as CT scan axial positions, variations of pneumonia etc. are all considered to be the noise for the model. In general, this can be attributed to the way conditional generative models e.g. ACGAN, CAGlow function. There can be numerous variations of synthetic samples that can be created using our model, keeping the conditional information same, hence a potential negative societal impact of our work can be misuse of synthetic medical information to spread misinformation.  
\section{Future Work}
With appropriate changes in the network and the training mechanism, our method can be generalized for synthetic generation of other kinds of data, e.g. X-rays, natural language and time-series. One of the future works includes the performance study of healthcare ML models, e.g. out-of-distribution models for new disease detection with synthetic data augmentation. Another interesting line of research is synthetically realizing data corresponding to medical cases that would otherwise be ethically/practically hard to obtain. Research can also be done to incorporate domain knowledge into the deep network used to generate synthetic data. It can also be used alongside real-world applications ~\citep{zou2019wifi,zou2019machine,konstantakopoulos2019design,chen2021enforcing,periyakoil2021environmental,das2019novel,das2020occupants,liu2018personal,liu2019personal,donti2021machine,jin2018biscuit} where challenges such as class-imbalance and privacy is important and thus generating conditional synthetic data is helpful. We believe our proposed conditional synthetic data generation work will enable new avenues of research into synthetic realization of medical data and eventually robust models in healthcare AI, essential in the fight against future pandemics.

\bibliography{aaai22}
\section*{Appendix}

\balance
\section{Hyperparameters}\label{app:hyperparameters}
\textbf{Classifier}
\begin{itemize}
    \item Batch size: 64
    \item Optimizer: AdamW optimizer
    \item Learning rate: $1e-5$
    \item Learning rate decay parameters: $0.99$, $0.998$, $0.999$, $0.999$, $0.9998$, $0.9998$ for classifiers trained on 100\% of the training set, 5\%, 1\%,  0.5\%, 50 samples, and 20 samples respectively. The decay parameter was set to 0.99 during epochs with presumptive labels during semi-supervised training.
    \item Weight decay rate: $1e-7$
    \item Beta parameters: $(0.9, 0.999)$
\end{itemize}
\textbf{Conditional Generative Flow}
\begin{itemize}
    \item Batch size: 320 across 4 GPUs
    \item Optimizer: AdamW
    \item Learning rate: $5e-4$
    \item Learning rate decay: It had a warm-up period of 10 epochs and was decayed on an exponential schedule with decay parameter $0.99$.
    \item Weight decay: $1e-6$
    \item Beta parameters: $(0.5, 0.999)$
    \item Temperature for Gaussian Noise Sampling: $0.9$
\end{itemize}

\section{Network Architecture}\label{app:network}
\textbf{Classifier}\\
Our classifier network is based on COVIDNet, by \citet{Wang2020covidnet}. It is composed of lightweight projection-expansion-projection-extension (PEPX) modules. The PEPX modules consist of $1\times 1$ convolutions for first stage projection that projects input features to a lower dimension, $1\times 1$  to expand the features to a higher dimension different than that of the input features, a depth-wise representation of features to learn spatial characteristics with $3\times 3$ convolutions, $1\times 1$ convolutions to project features back to a lower dimension and finally $1\times 1$ convolutions to extend the channel dimensionality to produce the final features. We take the dimension of the conditional input ($z$) to be $32$, and perform $l_2$-normalization on it before feeding it to the conditional generative flow. 

\textbf{Conditional Generative Flow}
We use a variant of Glow~\citep{kingma2018glow} model that features a reorganized flow step, designed to reduce the number of invertible $1\times 1$ convolutions, together with a fine-grained multi-scale architecture. Each coupling layer consists a $3\times 3$ convolution with ELU~\citep{clevert2015fast} non-linearity, a $1\times 1$ convolution, a channel-wise summation with a condition vector, a non-linearity, and a final $3\times 3$ convolution. The condition vector is obtained by taking the embedding of the image at the penultimate layer of our classifier and projecting it to the hidden dimension of the $1\times 1$ convolution layer.

We use a 4 level flow, with granularity factor $M=4$. The first and last levels consist of 8 flow steps, and the two internal levels each consist of a sequence of 3 blocks of 8 flow steps. The hidden dimension of the affine coupling layer at each level is $24,512,512,512$ in that order. 

\section{Computation}\label{app:computation}
We used 4 NVIDIA Tesla V100 GPUs from our internal compute clusters for the experiments.

\end{document}